\documentclass{bmvc2k}

\usepackage{enumitem}
\newenvironment{tight_itemize}{
\begin{itemize}[leftmargin=20pt]
  \setlength{\topsep}{0pt}
  \setlength{\itemsep}{0pt}
  \setlength{\parskip}{0pt}
  \setlength{\parsep}{0pt}
}{\end{itemize}}

\title{Stacked Dense U-Nets with Dual Transformers for Robust Face Alignment}

\addauthor{Jia Guo*}{https://github.com/deepinsight/insightface}{1}
\addauthor{Jiankang Deng*}{https://jiankangdeng.github.io/}{2}
\addauthor{Niannan Xue}{https://ibug.doc.ic.ac.uk/people/nxue}{2}
\addauthor{Stefanos Zafeiriou}{https://wp.doc.ic.ac.uk/szafeiri/}{2}

\addinstitution{
 InsightFace\\
 Shanghai, China
}

\addinstitution{
 IBUG\\
 Imperial College London\\
 London, UK
}

\runninghead{Jia Guo, Jiankang Deng}{Stacked Dense U-Nets with Dual Transformers}

\def\etal{\emph{et al}\bmvaOneDot}

\begin{document}

\maketitle

\begin{abstract} 
Facial landmark localisation in images captured in-the-wild is an important and challenging problem. The current state-of-the-art revolves around certain kinds of Deep Convolutional Neural Networks (DCNNs) such as stacked U-Nets and Hourglass networks. In this work, we innovatively propose stacked dense U-Nets for this task. We design a novel scale aggregation network topology structure and a channel aggregation building block to improve the model's capacity without sacrificing the computational complexity and model size. With the assistance of deformable convolutions inside the stacked dense U-Nets and coherent loss for outside data transformation, our model obtains the ability to be spatially invariant to arbitrary input face images. Extensive experiments on many in-the-wild datasets, validate the robustness of the proposed method under extreme poses, exaggerated expressions and heavy occlusions. Finally, we show that accurate 3D face alignment can assist pose-invariant face recognition where we achieve a new state-of-the-art accuracy on CFP-FP ($98.514\%$).
\end{abstract}

\section{Introduction}

Facial landmark localisation~\cite{sagonas2016300,stefanos2017menpo,stefanos20173Dmenpo,deng2016m,yang2015facial,yang2017stacked,liu2017adaptive,liu2016dual,deng2017joint} in unconstrained recording conditions has recently received considerable attention due to wide applications such as human-computer interaction, video surveillance and entertainment. 2D and 3D \footnote{In this paper, the 3D facial landmarks refer to the 2D projections of the real-world 3D landmarks, which can preserve face structure and semantic consistency across extreme pose variations.} in-the-wild face alignments are very challenging as facial appearance can change dramatically due to extreme poses, exaggerated expressions and heavy occlusions.

The current state-of-the-art 2D face alignment benchmarks~\cite{sagonas2016300,stefanos2017menpo} revolve around applying fully-convolutional neural networks to predict a set of landmark heatmaps, where for a given heatmap, the network predicts the probability of a landmark's presence at each and every pixel. Since the heatmap prediction for face alignment is essentially a dense regression problem, (1) rich features representations that span resolutions from low to high, and (2) skip connections that preserve spatial information at each resolution, are extensively investigated to combine multi-scale representations to improve inference of where and what~\cite{ronneberger2015u,lin2017feature,newell2016stacked,shrivastava2016beyond}. In fact, the most recent state-of-the-art performance in 2D face alignment has been held for a while~\cite{yang2017stacked,stefanos2017menpo} and is also believed to be saturated~\cite{bulat2017binarized,bulat2017far} by the stacked Hourglass models~\cite{newell2016stacked}, which repeat resolution-preserved bottom-up and top-down processing in conjunction with intermediate supervision.

Although lateral connections can consolidate multi-scale feature representations in Hourglass, these connections are shallow themselves due to simple one-step operations. Deep layer aggregation (DLA)~\cite{yu2017deep} augments shallow lateral connections with deeper aggregations to better fuse information across layers. We further add the down-sampling paths for the aggregation nodes in DLA and create a new Scale Aggregation Topology (SAT) for network design. Following the same insight in the network topology structure, we propose a Channel Aggregation Block (CAB). The decreasing channel in CAB helps to increase contextual modelling, which incorporates global landmark relationships and increases robustness when local observation is blurred. By combining SAT and CAB, we create the network structure designated dense U-Net. Nevertheless, the computation complexity and model size of the proposed dense U-Net dramatically increases and there is optimisation difficulty during model training especially when the training data is limited. Therefore, we further simplify the dense U-net by removing one down-sampling step as well as substituting some normal convolutions with deep-wise separable convolutions and direct lateral connections. Finally, the simplified dense U-net maintains similar computational complexity and model size as Hourglass, but significantly improves the model's capacity.

\begin{figure}[t]
\begin{center}
\includegraphics[width=1\columnwidth]{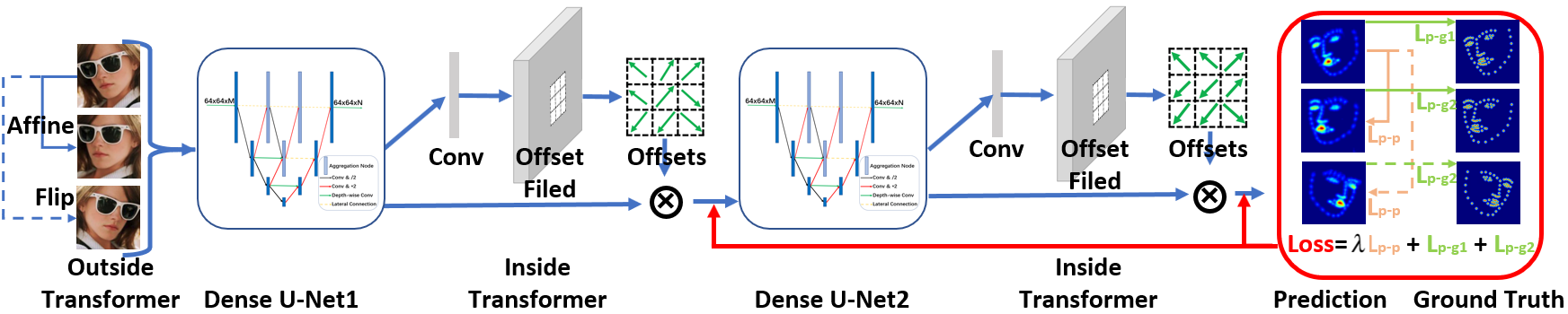}\\
\end{center}
\vspace{-2mm}
\caption{Stacked dense U-Nets with dual transformers for robust facial landmark localisation. We stack two dense U-Nets, each followed by a deformable convolution layer, as the network backbone. The input of the network is one face image together with its affine or flip transformed counterpart. The loss includes heatmap discrepancy between the prediction and ground truth as well as two predictions before and after transformation.}
\label{pic:framework}
\vspace{-2mm}
\end{figure}

Even though stacked dense U-Nets have a high capacity to predict the facial landmark heatmaps, they are still limited by the lack of ability to be spatially invariant to the input face images. Generally, the capability of modelling geometric transformations comes from deeper network design for transformation-invariant feature learning and extensive data augmentation. For transformation-invariant feature learning, Spatial Transform Networks (STN)~\cite{jaderberg2015spatial} is the first work to learn spatial transformation from data by warping the feature map via a global parametric transformation. However, such warping is expensive due to additional calculation on explicit parameter estimation. By contrast, deformable convolution~\cite{dai2017deformable} replaces the global parametric transformation and feature warping with a local and dense spatial sampling by additional offsets learning, thus introduces an extremely light-weight spatial transformer. For data augmentation, Honari~\etal~\cite{honari2017improving} have explored a semi-supervised learning technique for face alignment based on having a model predict equivariant landmarks with respect to transformations applied to the image. Similar idea can be found in~\cite{yang2015mirror}, where mirror-ability, the ability of a model to produce symmetric results in mirrored images, is explored to improve face alignment. Inspired by these works, we innovatively introduce dual transformers into the stacked dense U-Nets. As illustrated in Fig.~\ref{pic:framework}, inside the network, we employ deformable convolution to enhance transformation-invariant feature learning. Outside the network, we design a coherent loss for arbitrary transformed inputs, enforcing the model's prediction to be consistent with different transformations that are applied to the image. With the joint assistance of deformable convolution and coherent loss, our model obtains the ability to be spatially invariant to the arbitrary input face images.

In conclusion, our major contributions can be summarised as follows:
\vspace{-0.2cm}
\begin{tight_itemize}
\item We propose a novel scale aggregation network topological structure and a channel aggregation building block to improve the model's capacity without obviously increasing computational complexity and model size.
\item With the joint assistance of a deformable convolution inside the stacked dense U-Nets and coherent loss for outside data transformation, our model obtains the ability to be spatially invariant to the arbitrary input face images. 
\item The proposed method creates new state-of-the-art results on five in-the-wild face alignment benchmarks, IBUG~\cite{sagonas2016300}, COFW~\cite{burgos2013robust,ghiasi2015occlusion}, 300W-test~\cite{sagonas2016300}, Menpo2D-test~\cite{stefanos2017menpo} and AFLW2000-3D~\cite{zhu2016face}.
\item Assisted by the proposed 3D face alignment model, we make a breakthrough in the pose-invariant face recognition with the verification accuracy at $98.514\%$ on CFP-FP~\cite{sengupta2016frontal}.
\end{tight_itemize}

\section{Dense U-Net}


\subsection{Scale Aggregation Topology}

\begin{figure*}[ht!]
\centering
\subfigure[U-Net]{
\label{fig:subfig:unet}  
\includegraphics[height=0.22\textwidth]{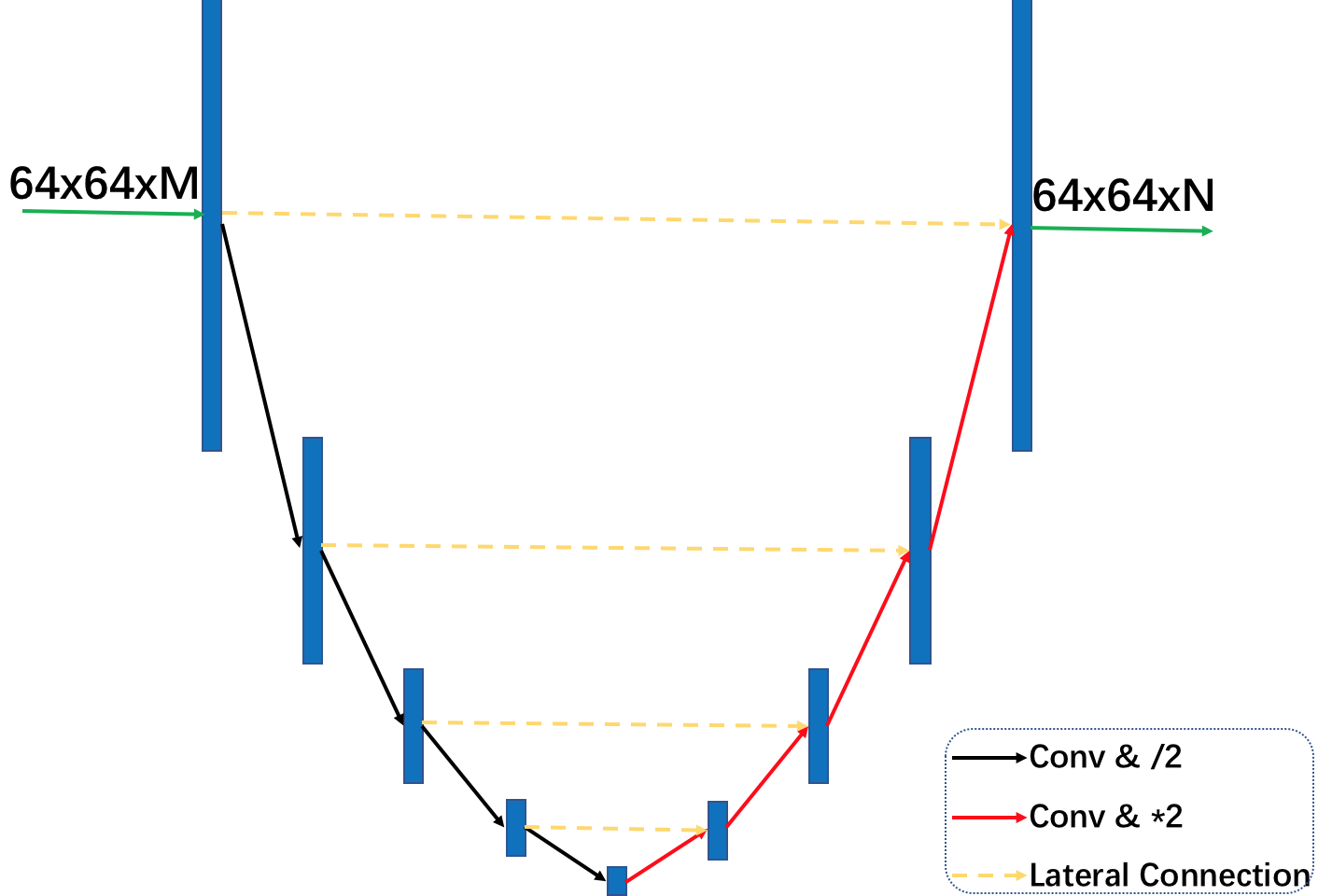}}
\subfigure[Hourglass]{
\label{fig:subfig:hourglass}  
\includegraphics[height=0.22\textwidth]{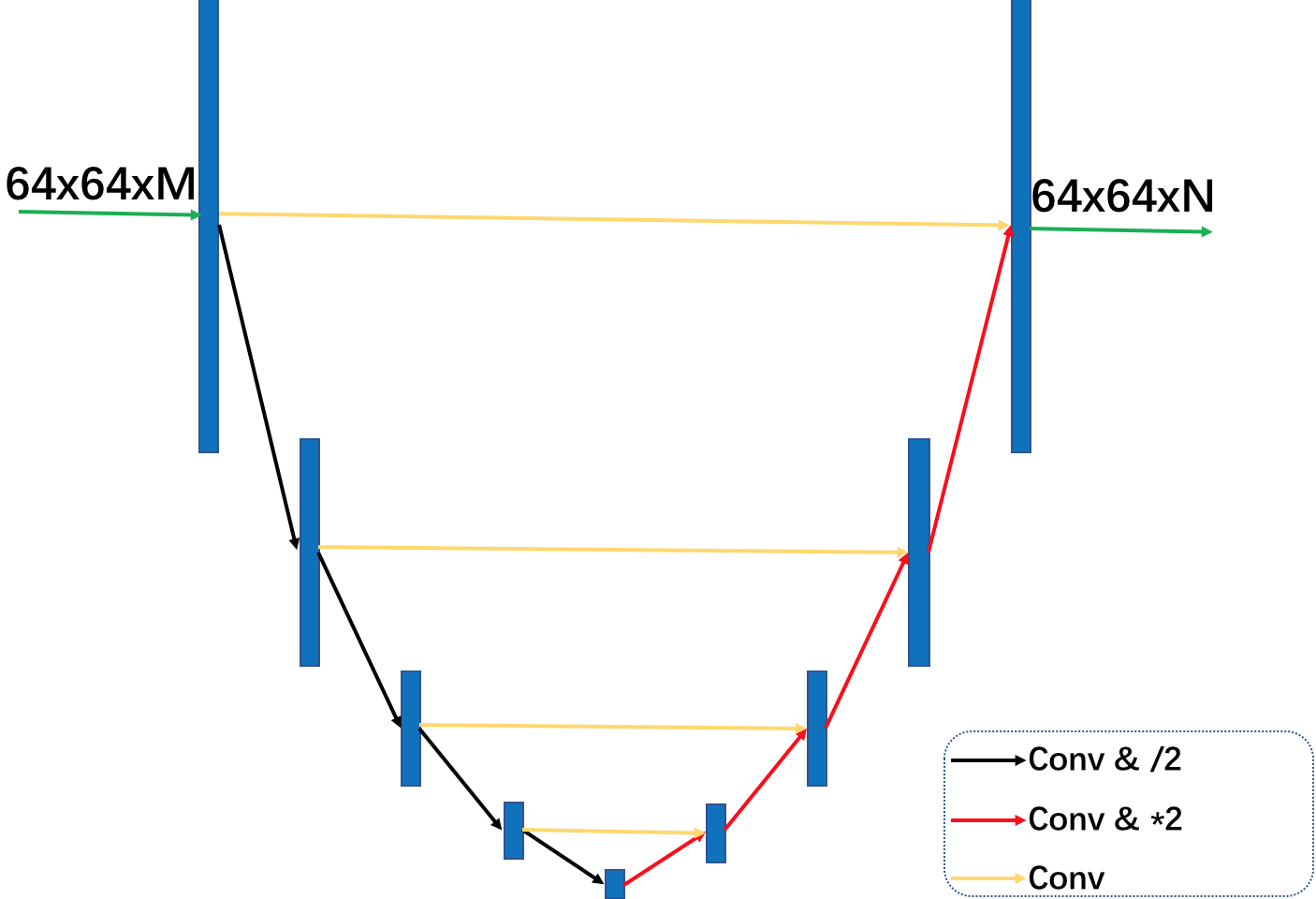}}
\subfigure[DLA]{
\label{fig:subfig:dla}  
\includegraphics[height=0.22\textwidth]{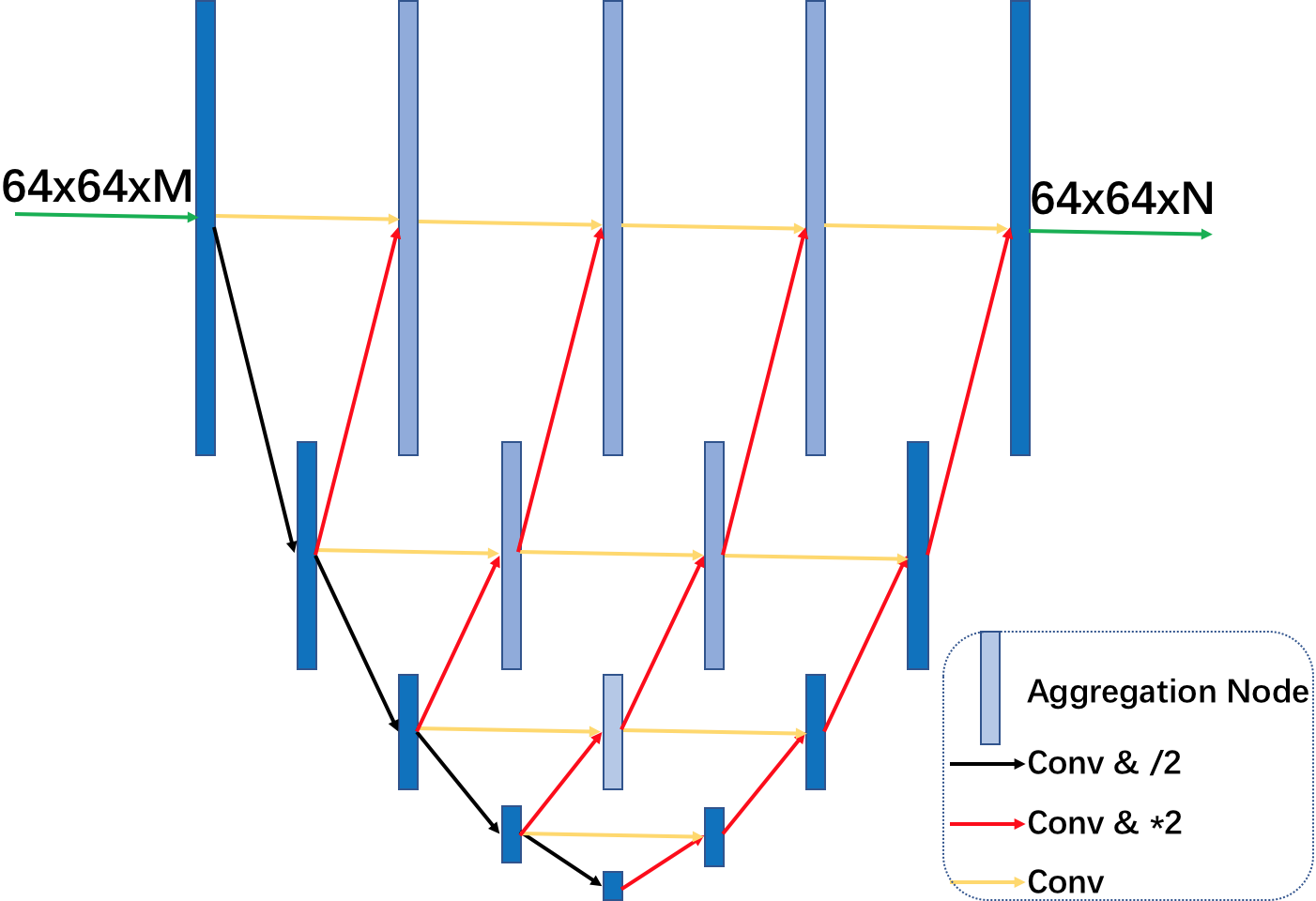}}
\subfigure[SAT (I)]{
\label{fig:subfig:denseunet}  
\includegraphics[height=0.22\textwidth]{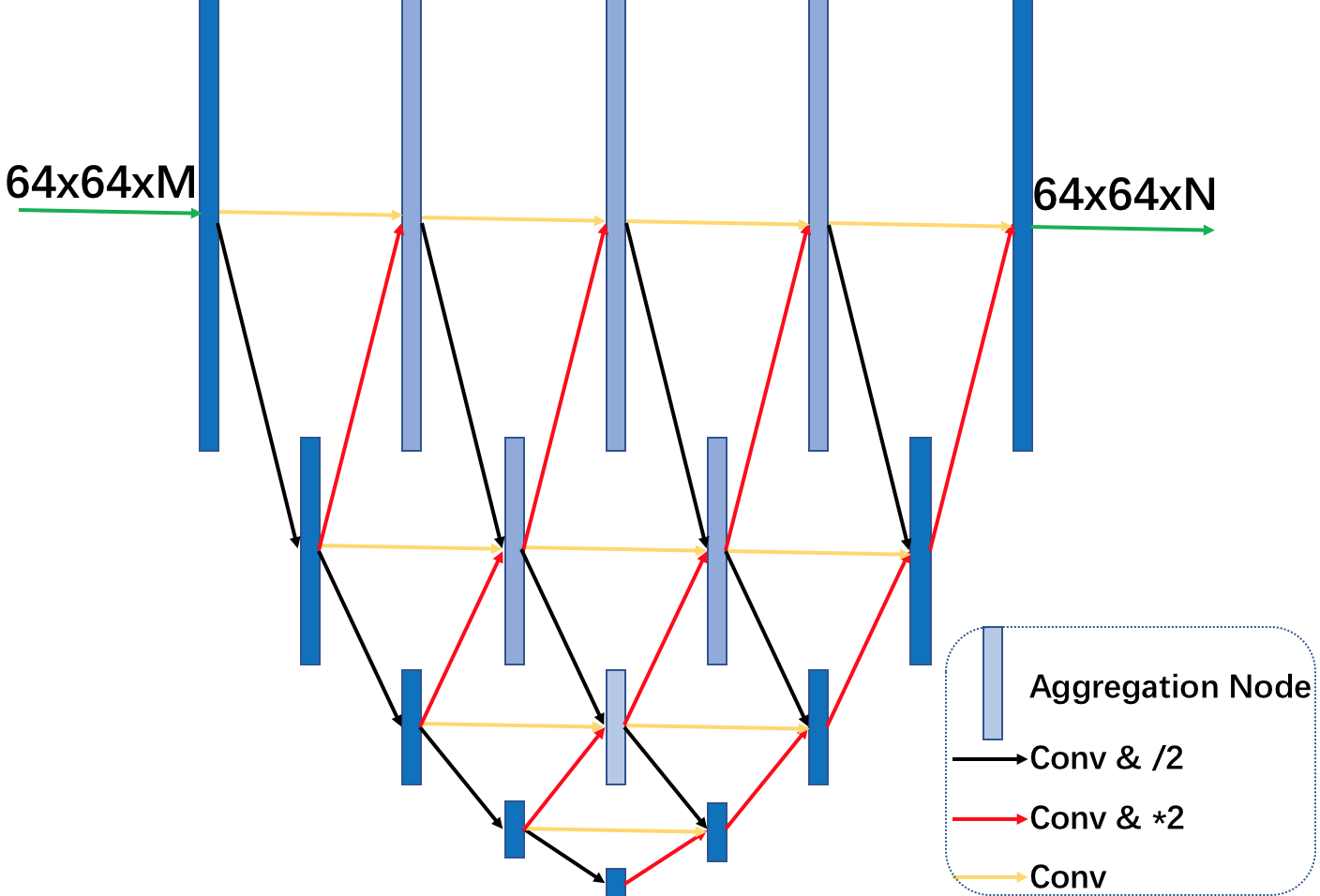}}
\subfigure[SAT (II)]{
\label{fig:subfig:sample3unet}  
\includegraphics[height=0.22\textwidth]{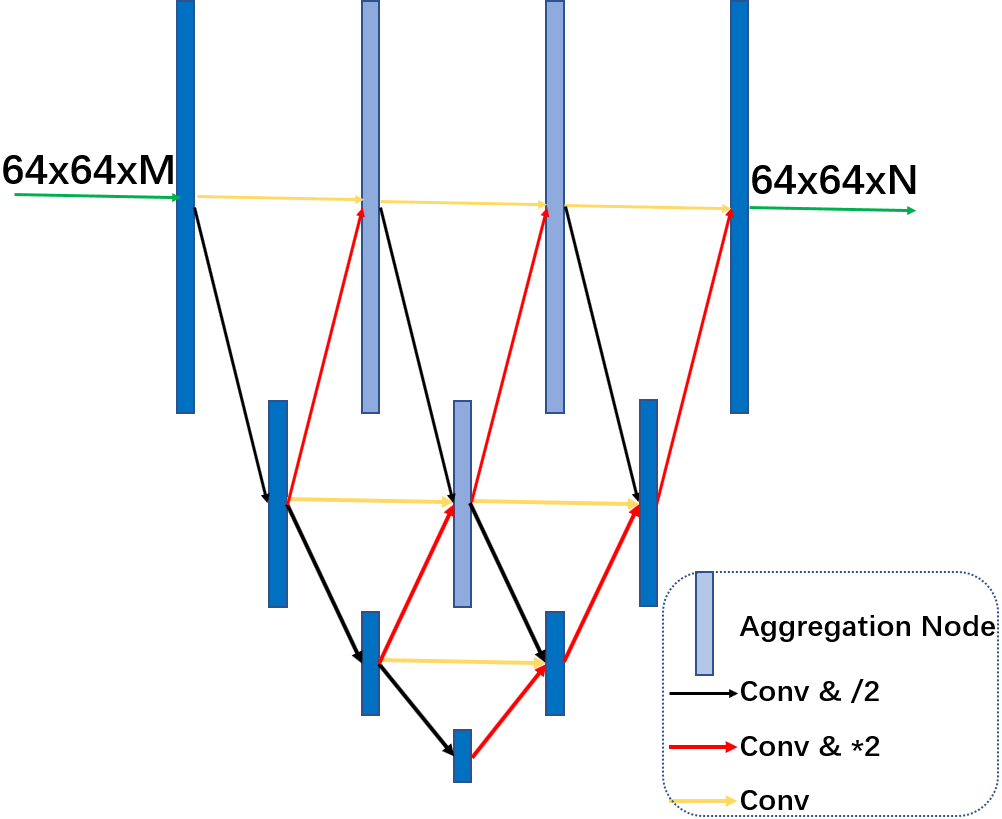}}
\subfigure[SAT (III)]{
\label{fig:subfig:simpledenseunet}  
\includegraphics[height=0.22\textwidth]{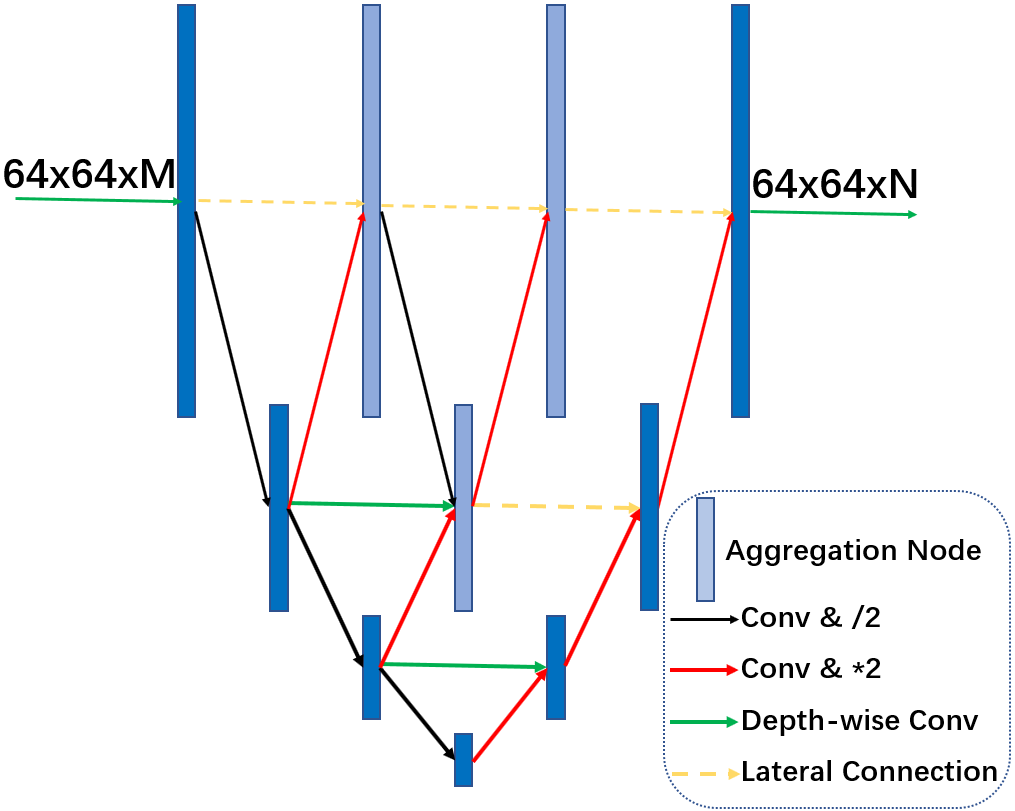}}
\caption{Different network topologies. SAT can capture local and global features and preserve spatial information by multi-scale information aggregation.}
\vspace{-4mm}
\label{fig:nettopology}
\end{figure*}

The essence of topology design for heatmap regression is to capture local and global features at different scales, while preserving the resolution information simultaneously. As illustrated in Fig.~\ref{fig:subfig:unet} and~\ref{fig:subfig:hourglass}, the topology of the U-Net~\cite{ronneberger2015u} and Hourglass~\cite{newell2016stacked} are both symmetric with four steps of pooling. At each down-sampling step, the network branches off the high resolution features, which are later combined into the corresponding up-sampling features. By using skip layers, U-Net and Hourglass can easily preserve spatial information at each resolution. Hourglass is similar to U-Net except for the extra convolutional layers within the lateral connections.


To improve the model's capacity, DLA (Fig.~\ref{fig:subfig:dla}) iteratively and hierarchically merges the feature hierarchy with additional aggregation nodes within the lateral connections. Inspired by DLA, we further propose a Scale Aggregation Topology (SAT) (Fig.~\ref{fig:subfig:denseunet}) by adding down-sampling inputs for aggregation nodes. The proposed SAT sets up a directed acyclic convolutional graph to aggregate multi-scale features for the pixel-wise heatmap prediction. However, the computation complexity and model size of SAT significantly builds up and the aggregation of three scale signals poses optimisation difficulty during model training especially when the training data is limited. To this end, we remove one step of pooling (Fig.~\ref{fig:subfig:sample3unet}), thus the lowest resolution is $8\times8$ pixels. In addition, we further remove some inner down-sampling aggregation paths and change some normal convolutions into depth-wise separable convolutions~\cite{howard2017mobilenets} and lateral connections~\cite{ronneberger2015u} as shown in Fig.~\ref{fig:subfig:simpledenseunet}. Finally, the simplified SAT maintains similar computational complexity and model size as Hourglass, but significantly improves the model's capacity.

\subsection{Channel Aggregation Block}

\begin{figure*}[h!]
\centering
\subfigure[Resnet]{
\label{fig:subfig:resnet}  
\includegraphics[height=0.26\textwidth]{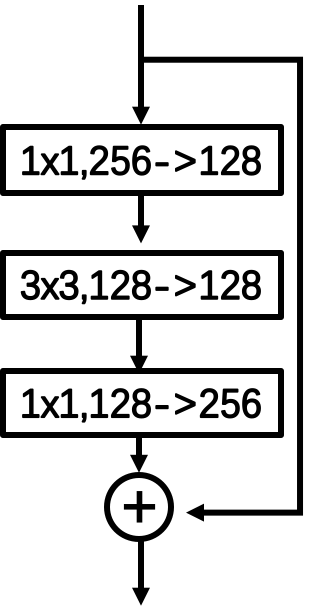}}
\subfigure[Inception-Resnet]{
\label{fig:subfig:inception}  
\includegraphics[height=0.26\textwidth]{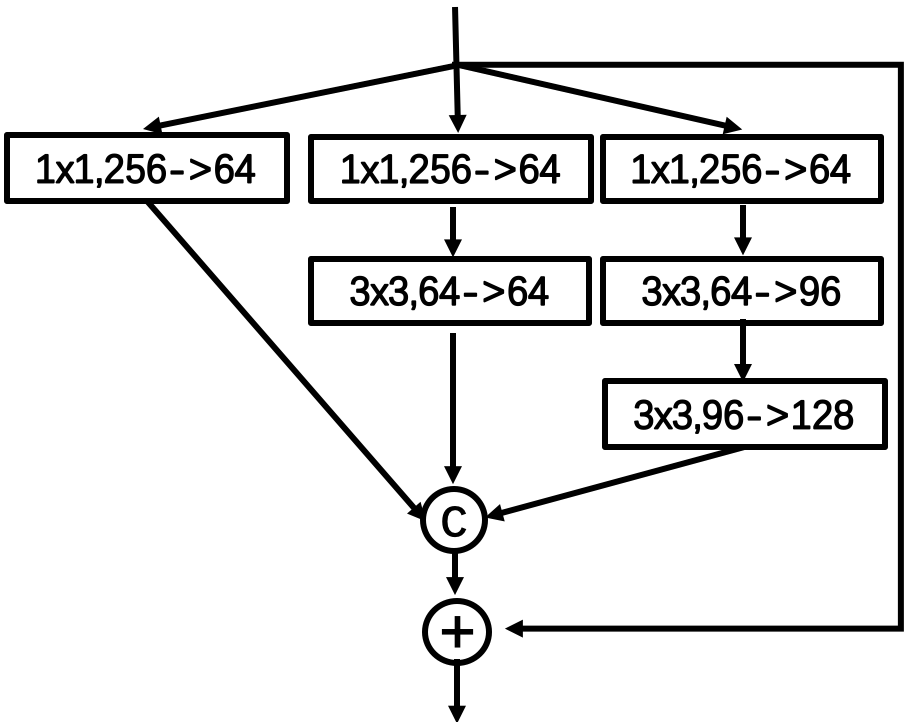}}
\subfigure[HPM~\cite{bulat2017binarized,bulat2017far}]{
\label{fig:subfig:HPM}  
\includegraphics[height=0.26\textwidth]{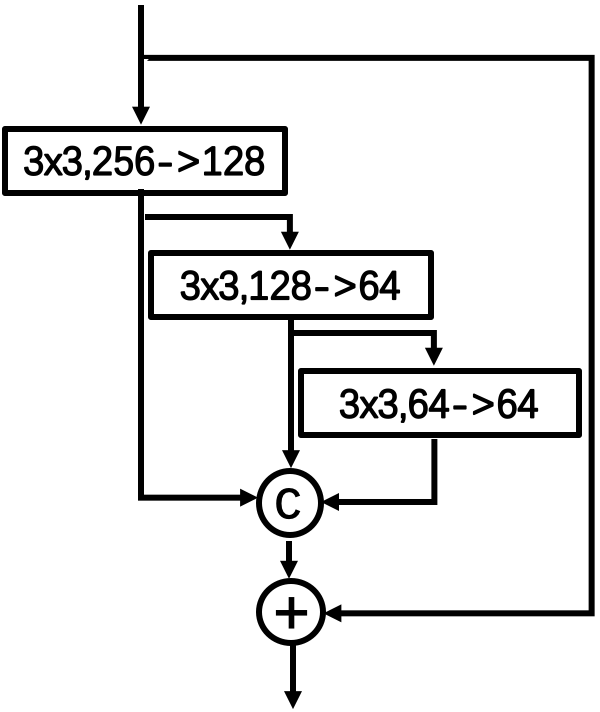}}
\subfigure[CAB]{
\label{fig:subfig:CAB}  
\includegraphics[height=0.42\textwidth]{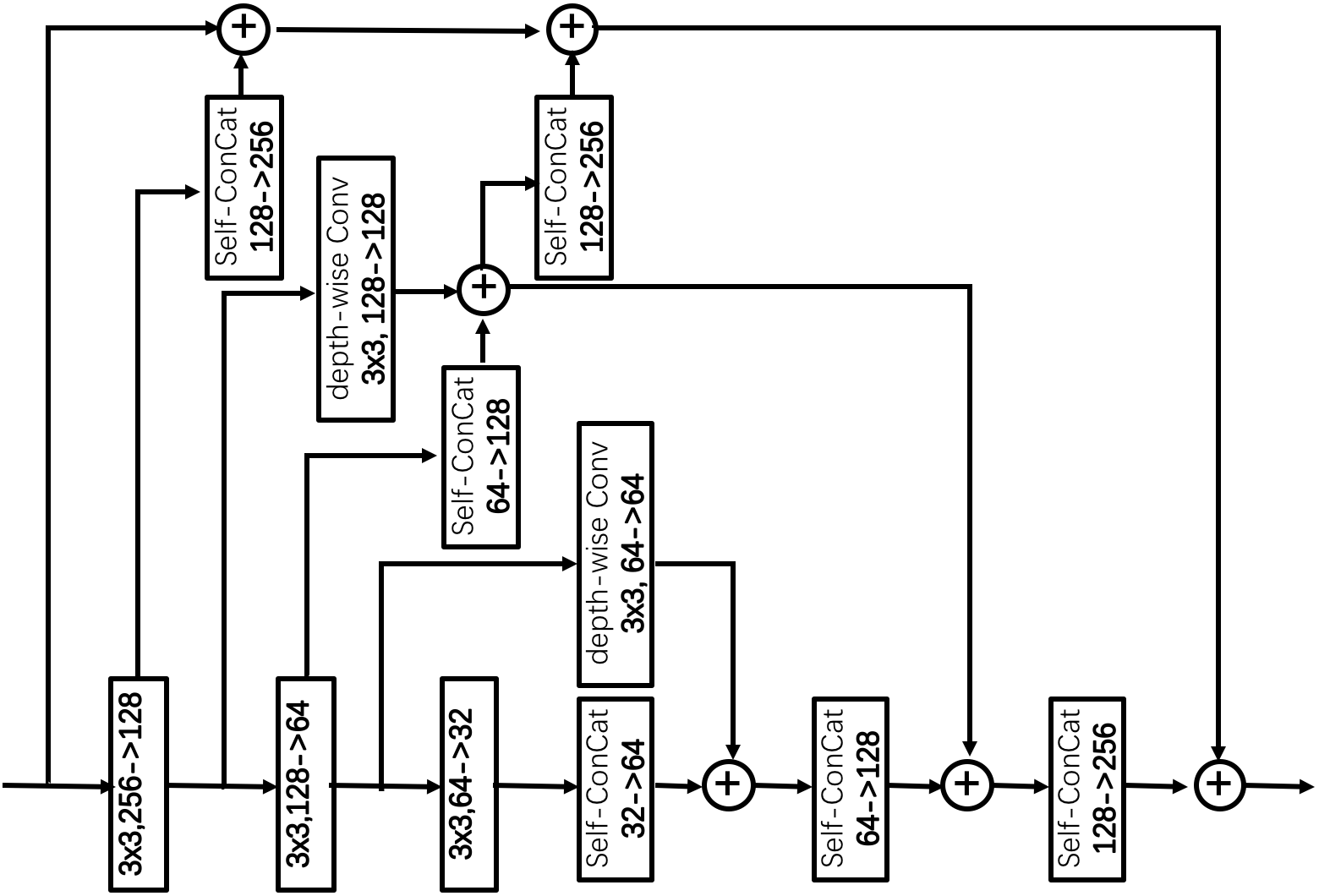}}
\caption{Different building blocks. CAB can enhance contextual modelling by channel compression and aggregation.}
\vspace{-4mm}
\label{fig:netblock}
\end{figure*}

The original Hourglass~\cite{newell2016stacked} employs the bottleneck residual block (Fig.~\ref{fig:subfig:resnet}). 
To improve the block's capacity, a parallel and multi-scale inception residual block is explored in~\cite{deng2018cascade} (Fig.~\ref{fig:subfig:inception}). Meanwhile, a novel hierarchical, parallel and multi-scale (HPM) residual block is extensively investigated in~\cite{bulat2017binarized,bulat2017far} (Fig.~\ref{fig:subfig:HPM}). 
For the building block design, we follow the same insight in the network topology and innovatively propose a Channel Aggregation Block (CAB). As shown in Fig.\ref{fig:subfig:CAB}, CAB is symmetric in channel while SAT is symmetric in scale. 
The input signals branch off before each channel decrease and converge back before each channel increase to maintain the channel information. Channel compression in the backbone can help contextual modelling~\cite{hu2017squeeze}, which incorporates channel-wise heatmap relationships and increases robustness when local observation is blurred. To control the computational complexity and compress the model size, depth-wise separable convolutions~\cite{howard2017mobilenets} and replication-based channel extensions are employed within CAB.

\section{Dual Transformer}

\subsection{Inside Transformer}

\begin{figure*}[h!]
\centering
\subfigure[Parameter Explicit Transformation]{
\label{fig:subfig:stn}  
\includegraphics[width=0.48\textwidth]{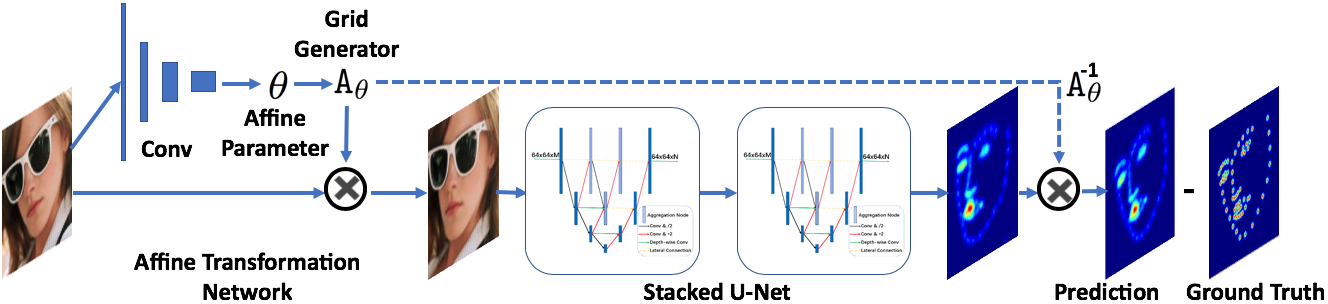}}
\subfigure[Parameter Implicit Transformation]{
\label{fig:subfig:deformable}  
\includegraphics[width=0.48\textwidth]{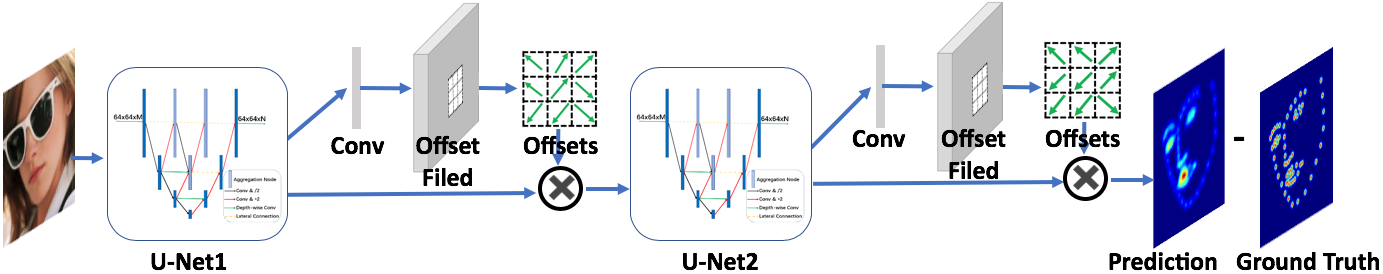}}
\caption{Inside transformer comparison: STN v.s. Deformable Convolution.}
\vspace{-4mm}
\label{fig:deformable}
\end{figure*}

We further improve the model's capacity by stacking two U-Nets end-to-end~\cite{newell2016stacked,bulat2017far}, feeding the output of the first U-Net as input into the second U-Net. Stacked U-Nets with intermediate supervision~\cite{newell2016stacked} provide a mechanism for repeated bottom-up, top-down inference allowing for re-evaluation and re-assessment of local heatmap predictions and global spatial configurations. However, stacked U-Nets still lack the transformation modelling capacity due to the fixed geometric structures. Here, we consider two different kinds of spatial transformers: parameter explicit transformation by STN~\cite{jaderberg2015spatial} and parameter implicit transformation by deformable convolution~\cite{dai2017deformable}. In Fig.~\ref{fig:subfig:stn}, we employ the STN to remove the discrepancy of rigid transformation (e.g. translation, scale and rotation) on the input face image, thus the following stacked U-Nets only need to focus on the non-rigid face transformation. Since the variance of the regression target is obviously decreased, the accuracy of face alignment can be easily improved. In Fig.~\ref{fig:subfig:deformable}, the application of deformable convolution behaves the similar way. Nonetheless, the deformable convolution augments the spatial sampling locations by learning additional offsets in a local and dense manner instead of adopting a parameter explicit transformation or warping. In this paper, we employ the deformable convolution as the inside transformer which is not only more flexible to model geometric face transformations but also has higher computation efficiency.

\subsection{Outside Transformer}

\begin{figure*}[h!]
\centering
\subfigure[Affine Transformation]{
\label{fig:subfig:affineloss}  
\includegraphics[width=0.3\textwidth]{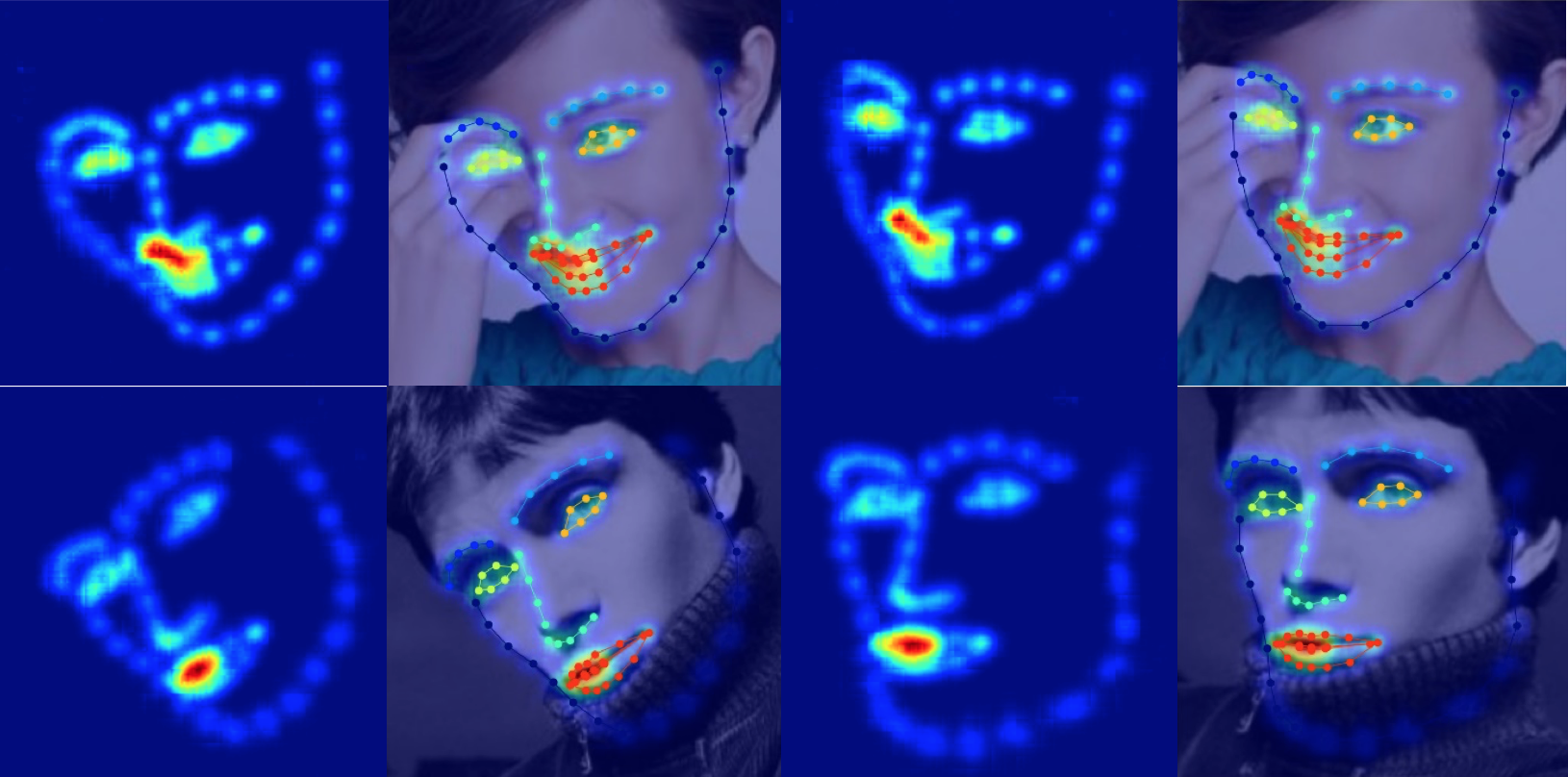}}
\subfigure[Flip]{
\label{fig:subfig:fliploss}  
\includegraphics[width=0.3\textwidth]{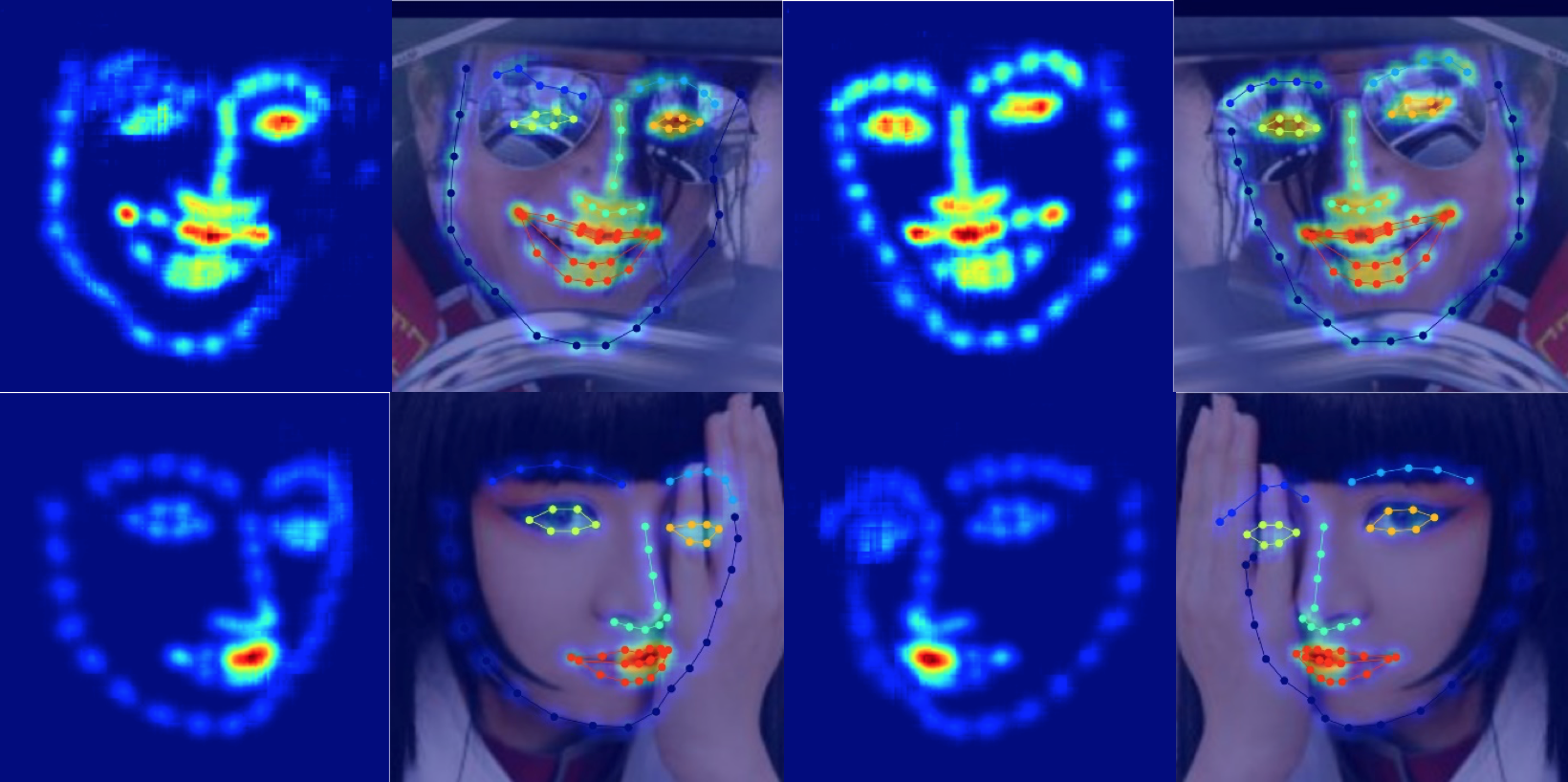}}
\caption{Heatmap incoherence under affine and flip transformation applied on the input images.}
\vspace{-4mm}
\label{fig:loss}
\end{figure*}

During training, data augmentation by random affine transformation on the input images is widely used to enhance the transformation modelling capacity. Nevertheless, the output heatmaps are not always coherent when there is affine or flip transformation on the input images~\cite{honari2017improving,yang2015mirror}. As illustrated in Fig.~\ref{fig:subfig:affineloss} and~\ref{fig:subfig:fliploss}, there are some obvious local differences between the heatmaps predicted from the original and transformed face images. 

We explore an outside transformer with an additional loss constraint, which encourages the regression network to output coherent landmarks when there are rotation, scale, translation and flip transformations applied to the images. More specifically, we transform an image during training and enforce the model to produce landmarks that are similarly transformed. Our model is trained end-to-end to minimise the following loss function
\begin{equation}
\small
L = \frac{1}{N} \sum_{n=1}^{N} (\lambda \underbrace{\| H_n(T \odot I) - T \odot H_n(I) \|_{2}^{2}}_{L_{p-p}} + \underbrace{\| H_n(I)-G_n(I)\|_{2}^{2}}_{L_{p-g1}}+ \underbrace{\| H_n(T \odot I) - T \odot G_n(I)\|_{2}^{2}}_{L_{p-g2}}),  
\label{eq:loss}
\vspace{-2mm}
\end{equation}
where $N$ is the landmark number, $I$ is the input image, $G(I)$ is the ground truth, $H(I)$ is the predicted heatmaps, $T$ is the affine or flip transformation, and $\lambda$ is the weight to balance two losses (in Fig.~\ref{pic:framework}): (1) the difference between the prediction and ground truth; and (2) the difference between two predictions before and after transformation.



\section{Experiments}

\subsection{Data}


For the training of 2D face alignment, we collate the training sets of the 300W challenge~\cite{sagonas2016300} and the Menpo2D challenge~\cite{stefanos2017menpo}. The {\em 300W-train} dataset consists of the LFPW, Helen and AFW datasets. The {\em Menpo2D-train} dataset consists of 5,658 semi-frontal face images, which are selected from FDDB and ALFW.
Hence, a total of 9,360 face images are used to train the 2D68 face alignment model.
We extensively test the proposed 2D face alignment method on four image datasets: 
the {\em IBUG} dataset (135 images)~\cite{sagonas2016300}, the {\em COFW} dataset (507)~\cite{burgos2013robust,ghiasi2015occlusion}, the {\em 300W-test} dataset (600)~\cite{sagonas2016300}, and the {\em Menpo2D-test} dataset (5,535)~\cite{stefanos2017menpo}.

For the training of 3D face alignment, we utilise the {\em 300W-LP} dataset~\cite{zhu2016face}, which contains 61,225 synthetic face
images. The {\em 300W-LP} is generated by profiling and rendering the faces of 300-W~\cite{sagonas2016300} into larger poses (ranging from $−90^\circ$ to $90^\circ$). We test the proposed 3D face alignment method on the {\em AFLW2000-3D} dataset (2,000)~\cite{zhu2016face}.

\subsection{Training Details}

Each face region is cropped and scaled to $128\times128$ pixels based on the face boxes~\cite{zhang2016joint}. We augment the ground truth image with a random combination of horizontal flip, rotation (+/- 40 degrees), and scaling (0.8 - 1.2). The network starts with a $3\times3$ convolutional layer, followed by a residual module and a round of max pooling to bring the resolution down from 128 to 64, as it could reduce GPU memory usage while preserving alignment accuracy. The network is trained using MXNet with Nadam optimiser, an initial learning rate of $2.5^{-4}$, a batch size of 40, and 30k learning steps. 
We drop the learning rate by a ratio of $0.2$ after 16k and 24k iterations. Each training step paralleled on two NVIDIA GTX Titan X (Pascal) takes 1.233s. Although the Mean Squared Error (MSE) pixel-wise loss is given in Eq.~\ref{eq:loss}, in practice we find the Sigmoid Cross-Entropy (CE) pixel-wise loss~\cite{bulat2017binarized} outperforms the MSE loss for $L_{p-g}$. Therefore, we employ the CE loss for $L_{p-g}$ and the MSE loss for $L_{p-p}$, respectively. $\lambda$ is empirically set as $0.001$ to guarantee convergence.


\subsection{Ablation Experiments}

In Tab.~\ref{tab:ablation}, we compare the alignment accuracy on the most challenging datasets (IBUG and COFW) under different training settings. We denote each training strategy by ${topology}^{stack}$-$block$ ($\downarrow \times$ $down$-$sampling\ steps$, $4$ by default). 

\begin{table}
\begin{center}
\begin{tabular}{c|c|c|c|c}
\hline
Method                                            & IBUG(\%) & COFW(\%) & Size (mb) & Time (ms)\\
\hline
$Hourglass^1$-$Resnet$~\cite{newell2016stacked}     & 7.32  & 6.26  & 13 & 26 \\
$Hourglass^2$-$Resnet$~\cite{newell2016stacked}     & 7.22  & 6.18  & 26 & 49 \\
\hline
$Hourglass^2$-$Inception$-$Resnet$                    & 7.07  & 6.08  & 38 & 57 \\
$Hourglass^2$-$HPM$~\cite{bulat2017far}             & 6.98  & 5.81  & 48 & 47 \\
$Hourglass^2$-$CAB$                                 & 6.93  & 5.77  & 46 & 52 \\    
\hline
$Hourglass^2$-$HPM$ ($\downarrow \times 3$)~\cite{bulat2017far} & 6.95  & 5.82  & 38 & 39 \\ 
$Hourglass^2$-$CAB$ ($\downarrow \times 3$)                     & 6.91  & 5.78  & 37 & 41 \\
\hline
$U$-$Net^2$-$CAB$                        &  7.17  & 6.12  & 36   & 37\\
$Hourglass^2$-$CAB$                      &  6.93  & 5.77  & 46   & 52 \\    
$DLA^2$-$CAB$                            &  6.92  & 5.75  & 103  & 61\\
$SAT (I)$   $^2$-$CAB$                   &  7.05  & 5.91  & 131  & 63\\
$SAT (II)$  $^2$-$CAB$ ($\downarrow \times 3$)           & 7.02  & 5.89  & 83 & 47\\
$SAT (III)$ $^2$-$CAB$ ($\downarrow \times 3$)           & 6.88  & 5.74  & 38 & 41 \\ 
\hline
$Dense U$-$Net$ + STN                                                         & 6.81  & 5.70  & 116 & 49\\
$Dense U$-$Net$ + Inside Transformer                                          & 6.77  & 5.63  & 49 & 45 \\
\hline
$Dense U$-$Net$ + Outside Transformer                                         & 6.80  & 5.62  & 38 & 41\\
\hline
$Dense U$-$Net$ + Dual Transformer                                            & {\bf 6.73}  & {\bf 5.55}  & 49 & 45 \\
\hline
\end{tabular}
\end{center}
\vspace{-2mm}
\caption{Ablation study for different settings on the IBUG and COFW datasets. Performance is reported as the eye centre distance normalised mean error.}
\vspace{-4mm}
\label{tab:ablation}
\end{table}

From Tab.~\ref{tab:ablation}, we can draw the following conclusions: (1) Compared to a single Hourglass network, a stack of two Hourglass networks can significantly improve the alignment accuracy even though the model size and inference time have been doubled; (2) Building blocks, such as Inception-Resnet, HPM~\cite{bulat2017far} or CAB, can progressively improve the performance with similar model size and computation cost; (3) The performance gap is not apparent between three and four down-sampling steps, but three down-sampling steps can remarkably decrease the model size; (4) As the complexity of the network topology increases from U-Net, to Hourglass and to DLA, the localisation accuracy gradually raises; (5) Due to limited training data ($\sim 10k$) and the optimisation difficulty with SAT (I) (Fig.~\ref{fig:subfig:denseunet}), e.g. aggregation of three scale signals, the performance obviously drops. By removing some down-sampling paths, introducing depth-wise separable convolutions and applying direct lateral convolutions (Fig.~\ref{fig:subfig:simpledenseunet}), the performance bounces back and eventually surpasses Hourglass and DLA; (6) Deformable convolution outperforms STN even with much fewer parameters; (7) Outside transformer with coherent loss can also evidently reduce the alignment error; (8) The proposed stacked dense U-Nets with dual transformers transcend recent state-of-the-art methods on the IBUG ($7.02\%$ CVPR18~\cite{feng2017wing}) and COFW ($5.77\%$ CVPR18~\cite{kumar2018disentangling}) datasets without bells and whistles.

\subsection{2D and 3D Face Alignment Results}

\begin{figure*}[h!]
\centering
\subfigure[COFW]{
\label{fig:subfig:cedcofw}  
\includegraphics[width=0.30\textwidth]{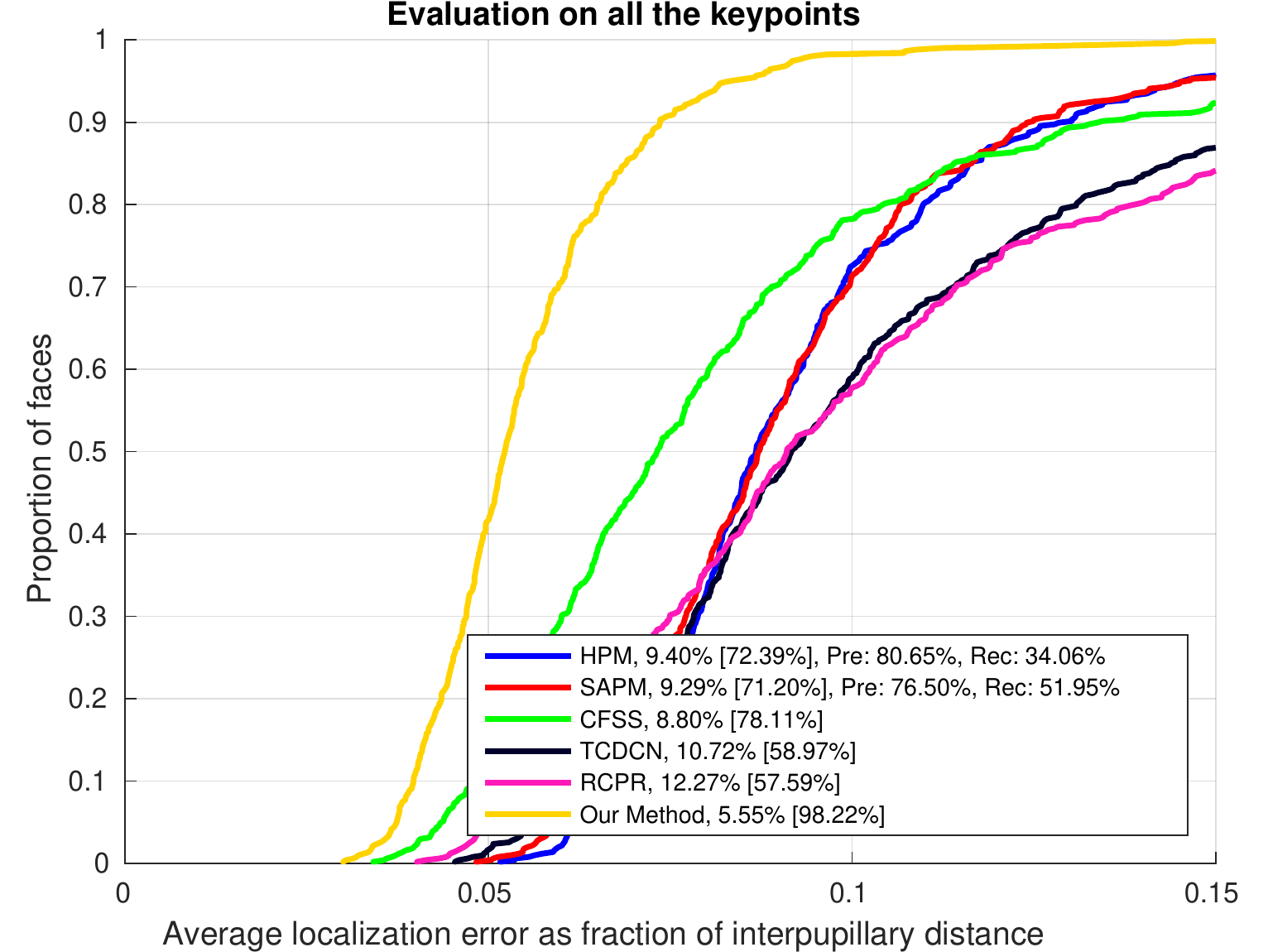}}
\subfigure[300W-test]{
\label{fig:subfig:ced300W}  
\includegraphics[width=0.30\textwidth]{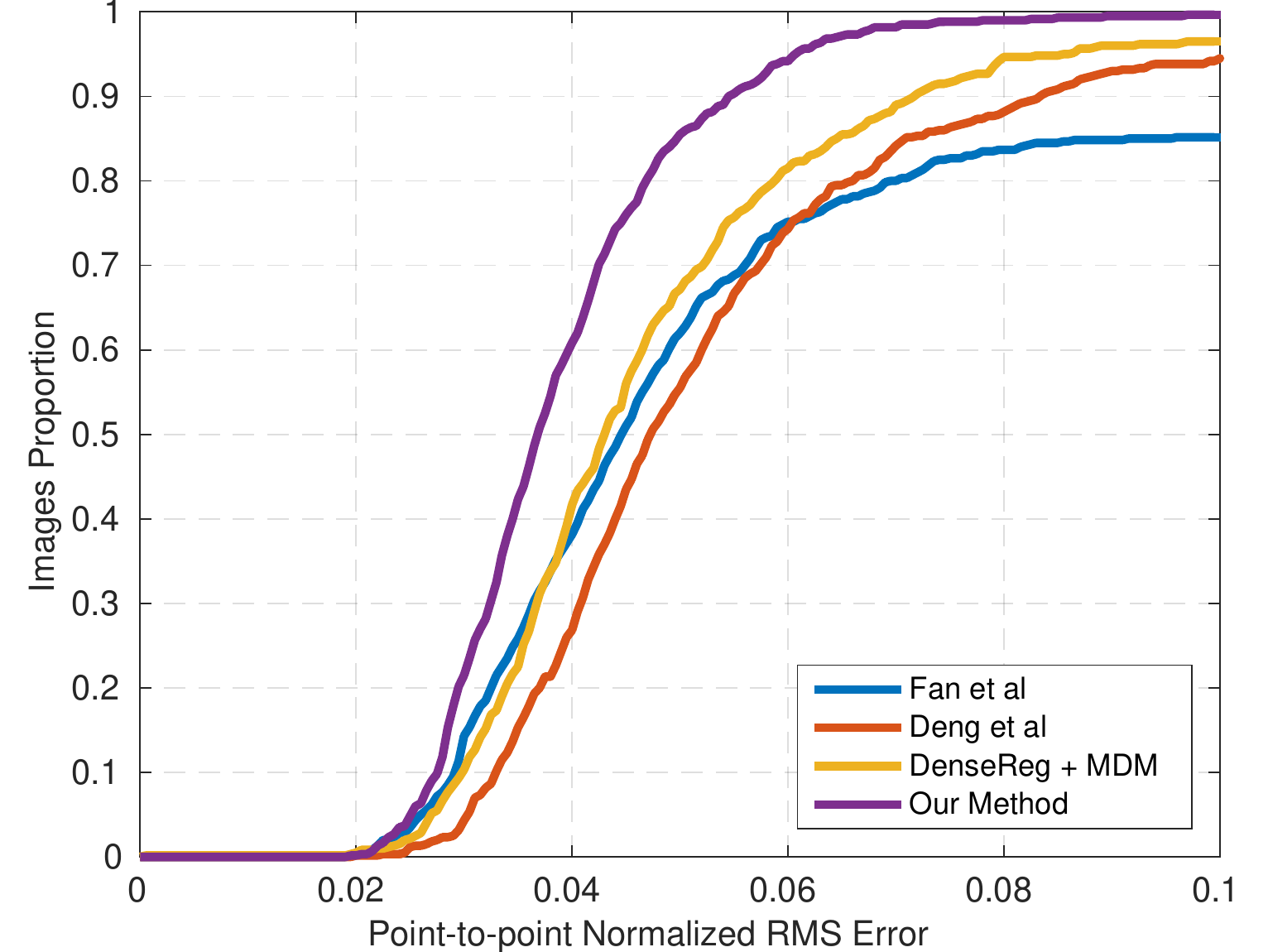}}
\subfigure[Menpo2D-test]{
\label{fig:subfig:cedMenpo}  
\includegraphics[width=0.32\textwidth]{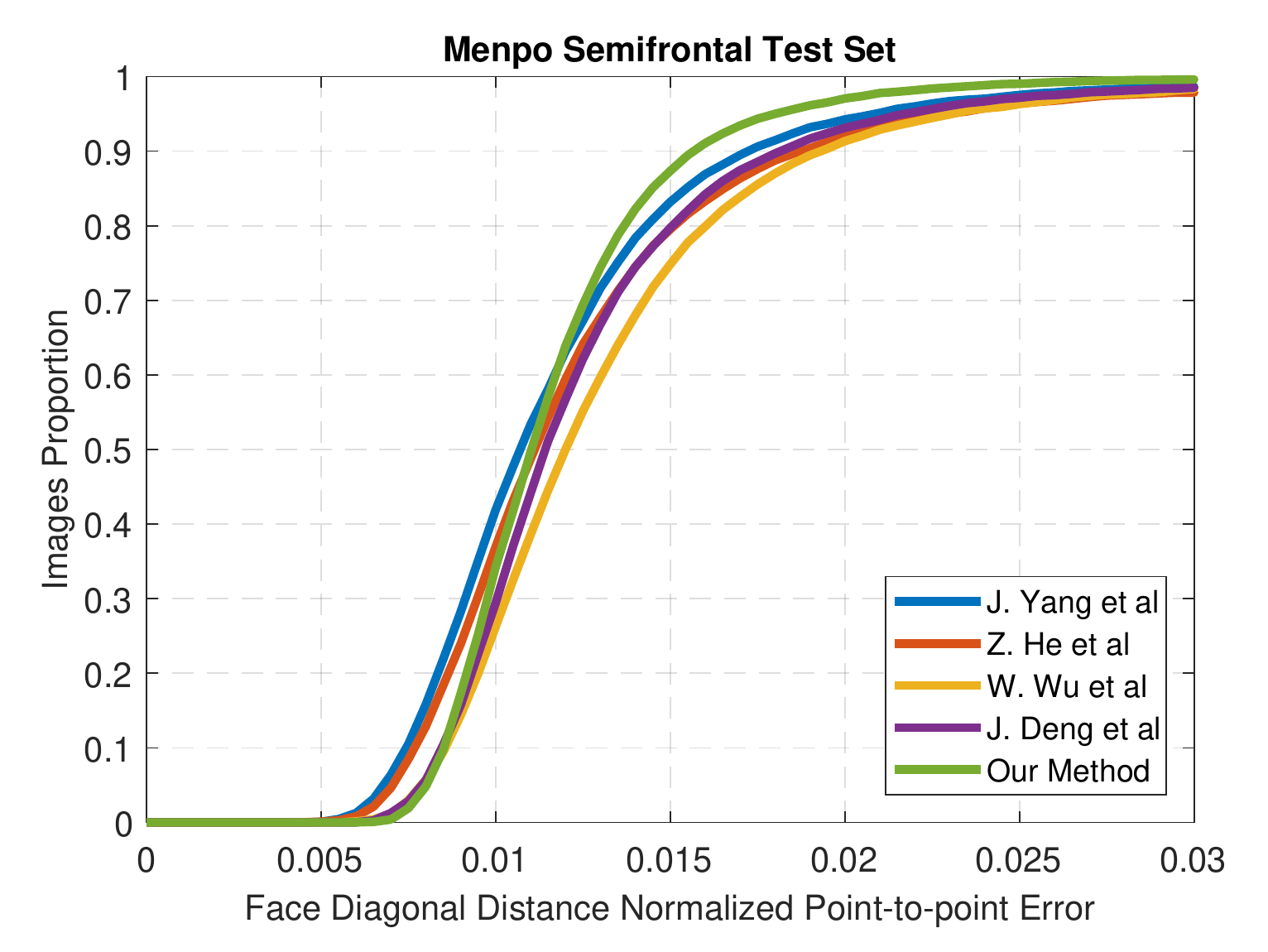}}
\caption{Landmark localisation results on the COFW, 300W-test, Menpo2D-test datasets. Performance is reported as mean error normalised by the eye centre distance (COFW), the out eye corner distance (300W-test), and the diagonal of the ground truth bounding box (Menpo2D-test), respectively.}
\vspace{-4mm}
\label{fig:ced}
\end{figure*}

\begin{figure*}[h!]
\centering
\vspace{-2mm}
\subfigure[IBUG]{
\label{fig:subfig:IBUG} 
\includegraphics[width=1\textwidth]{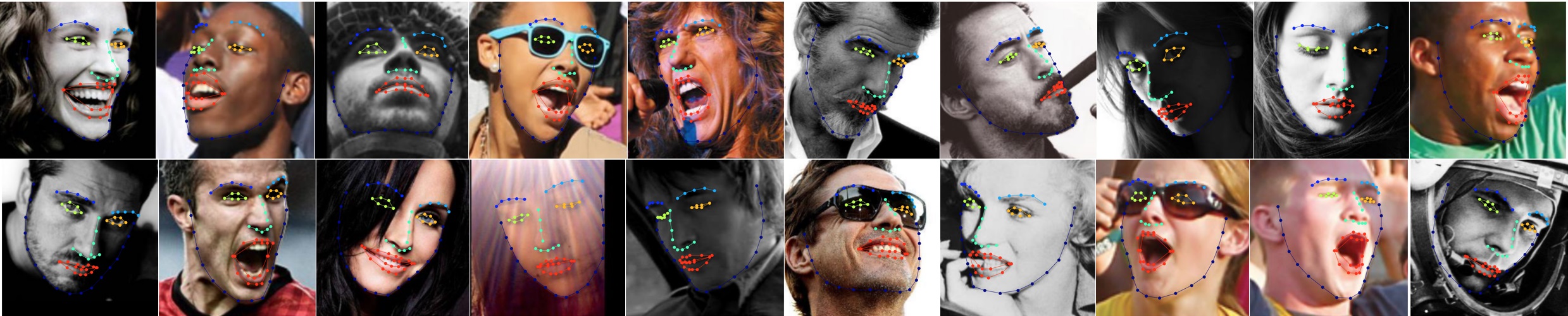}}
\vspace{-2mm}
\subfigure[COFW]{
\label{fig:subfig:COFW}  
\includegraphics[width=1\textwidth]{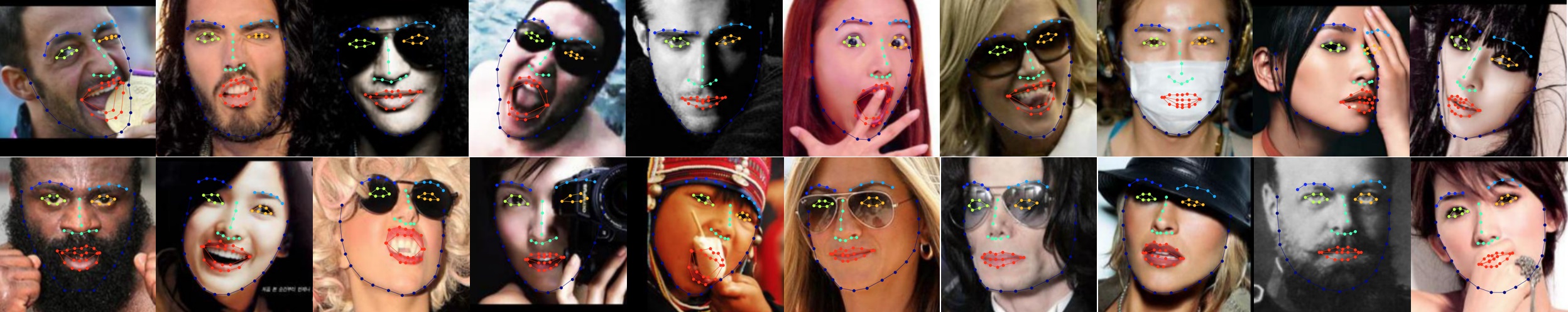}}
\vspace{-2mm}
\subfigure[300W-test]{
\label{fig:subfig:300wtest}  
\includegraphics[width=1\textwidth]{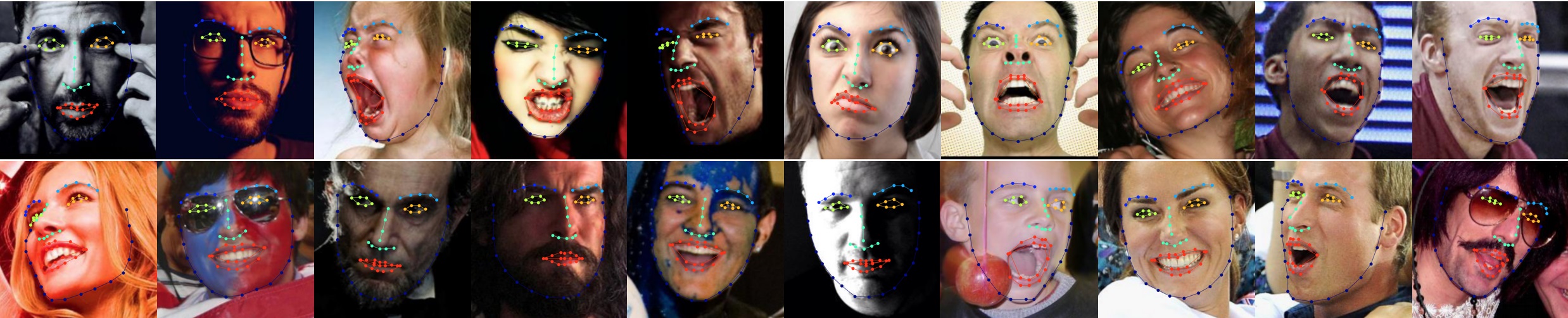}}
\vspace{-2mm}
\subfigure[Menpo2D-test]{
\label{fig:subfig:menpo2dtest}  
\includegraphics[width=1\textwidth]{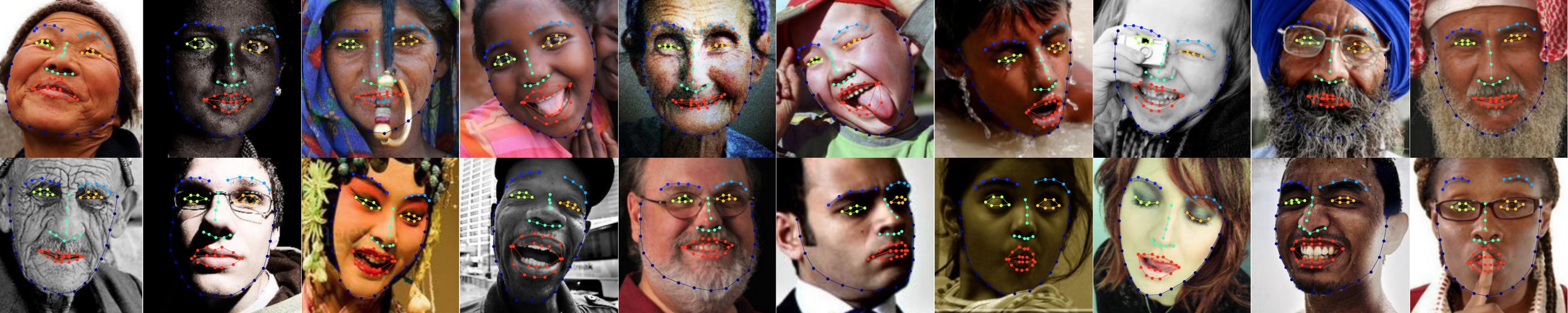}}
\vspace{-2mm}
\subfigure[AFLW2000-3D]{
\label{fig:subfig:aflw2000}  
\includegraphics[width=1\textwidth]{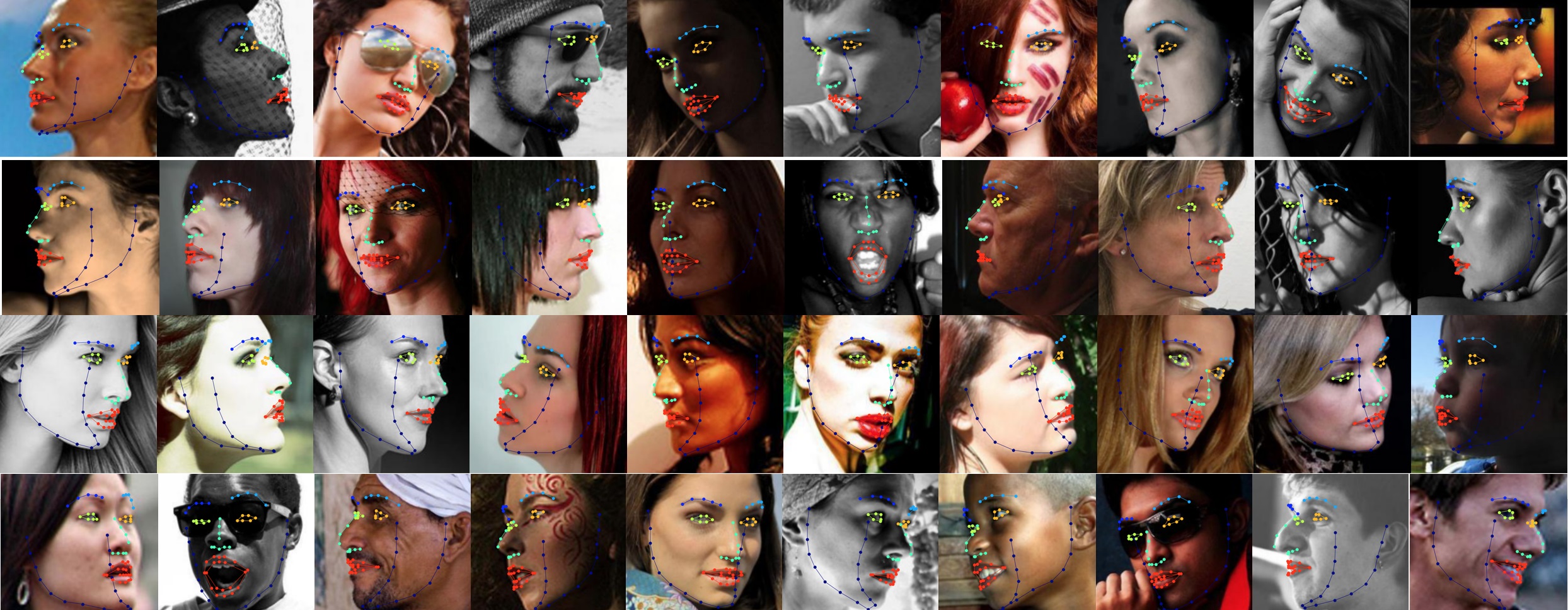}}
\vspace{-2mm}
\caption{Example results of 2D and 3D face alignment. The proposed method is robust under pose, expression, occlusion and illumination variations.}
\vspace{-2mm}
\label{fig:examples}
\end{figure*}

For 2D face alignment, we further report the Cumulative Error Distribution (CED) curves on three standard benchmarks, that is COFW, 300W-test, Menpo2D-test. On COFW~\cite{burgos2013robust} (Fig.~\ref{fig:subfig:cedcofw}), our method outperforms the baseline methods in the evaluation toolkit by a prominent margin. The normalised mean error of $5.55\%$ as well as the final success rate at $98.22\%$ are so impressive that the challenge of face alignment under occlusion is even no longer remarkable for our method. In Fig.~\ref{fig:subfig:COFW}, we give some fitting examples on COFW under heavy occlusions. Even by zooming in on these visualised results, we can hardly find any alignment flaw, which indicates that the proposed method can easily capture and consolidate local evidence and global context, and thus improve the model's robustness under occlusions.

On 300W-test~\cite{sagonas2016300}, we compare our method with leading results, such as Deng~\etal~\cite{deng2016m} and Fan~\etal~\cite{fan2016approaching}. Besides, we also compare with the state-of-the-art face alignment method ``DenseReg + MDM''~\cite{guler2016densereg}. Once again, our model surpasses those methods with ease. On Menpo2D-test~\cite{stefanos20173Dmenpo}, we send our alignment results to the organiser and get the CED curves with other best four entries of the competition~\cite{stefanos2017menpo}. As shown in Fig.~\ref{fig:subfig:cedMenpo}, we find our performance is inferior to the best entry~\cite{yang2017stacked} within the high accurate interval (NME $<1.2\%$) because our alignment model is initialised from MTCNN~\cite{zhang2016joint} which is less accurate and stable than the detectors applied in~\cite{yang2017stacked}. Nevertheless, our model gradually outperforms the best entry, which indicates that our model is more robust under hard cases, such as large pose variations, exaggerated expressions and heavy occlusions. 

\begin{table}[h!]
\begin{center}
\begin{tabular}{c|c|c|c|c|c|c}
\hline
Method  & ESR~\cite{cao2012face} & RCPR~\cite{burgos2013robust} & SDM~\cite{xiong2013supervised} & 3DDFA~\cite{zhu2016face} & HPM~\cite{bulat2017binarized} &  Our Method  \\
\hline
NME   & 7.99   & 7.80  & 6.12  & 4.94  & 3.26 & {\bf 3.07}\\
\hline
\end{tabular}
\end{center}
\vspace{-2mm}
\caption{3D alignment results on the AFLW2000-3D dataset. Performance is reported as the bounding box size normalised mean error~\cite{zhu2016face}.}
\vspace{-6mm}
\label{table:AFLW20003D}
\end{table}

For 3D face alignment, we compare our model on AFLW2000-3D~\cite{zhu2016face} with the most recent state-of-the-art method proposed by Bulat~\etal~\cite{bulat2017binarized}, which claimed that the problem of face alignment is almost solved with saturated performance. Nonetheless, our method further decreases the NME by $5.8\%$. In Fig.~\ref{fig:subfig:aflw2000}, we give some exemplary alignment results, which demonstrate successful 3D face alignment under extreme poses (ranging from $-90^\circ$ to $90^\circ$), accompanied by exaggerated expressions and heavy occlusions. 

\subsection{3D Face Alignment Improves Face Recognition}

Even though face alignment is not claimed to be essential for deep face recognition~\cite{schroff2015facenet,deng2017marginal,deng2017uv,deng2018arcface}, the face normalisation step is still widely applied in recent state-of-the-art recognition methods~\cite{deng2018arcface}. Following the setting of ArcFace~\cite{deng2018arcface}, we train recognition models under different face alignment methods on VGG2~\cite{cao2017vggface2}. As we can see from Tab.~\ref{table:facerecognition}, the verification performance on LFW~\cite{huang2007labeled} is comparatively close. However, on CFP-FP~\cite{sengupta2016frontal}, the proposed 3D alignment method obviously decreases the versification error by $48.24\%$ compared to the alignment method proposed in~\cite{zhang2016joint}. The significant improvement implies that accurate full-pose face alignment can hugely assist deep face recognition.

\begin{table}[h!]
\begin{center}
\begin{tabular}{c|c|c|c}
\hline
Testset       & No Alignment   & 2D-5 landmarks~\cite{zhang2016joint} & 3D-68 landmarks\\
\hline
LFW   &  99.63 & 99.78  & {\bf 99.80} \\
CFP-FP &  95.428 & 97.129 & {\bf 98.514} \\
\hline
\end{tabular}
\end{center}
\vspace{-2mm}
\caption{Face verification accuracy ($\%$) on the LFW and CFP-FP dataset (ArcFace, LResNet50E-IR@VGG2-LFW-CFP).}
\vspace{-6mm}
\label{table:facerecognition}
\end{table}

\section{Conclusion}

In this paper, we propose stacked dense U-Nets with dual transformers for robust 2D and 3D facial landmark localisation.
We introduce a novel network structure (Scale Aggregation Topology) and a new building block (Chanel Aggregation Block) to improve the model's capacity without sacrificing computational complexity and model size. With the assistance of deformable convolution and coherent loss, our model obtains the ability to be spatially invariant to the input face images. Extensive experiments on five challenging face alignment datasets demonstrate the robustness of the proposed alignment method. The additional face recognition experiment suggests that the proposed 3D face alignment can obviously improve pose-invariant face recognition.


\section{Acknowledgement}

J. Deng is supported by the President's Scholarship of Imperial College London. This work is also funded by the EPSRC project EP/N007743/1 (FACER2VM) and the European Community Horizon 2020 [H2020/2014-2020] under grant agreement no. 688520(TeSLA). We thank the NVIDIA Corporation for the GPU donations.

\bibliography{egbib}
\end{document}